\documentclass[a4paper,conference]{IEEEtran}

\usepackage{graphicx}
\usepackage[nohyperlinks,nolist]{acronym}
\usepackage[binary-units=true]{siunitx}
\usepackage{xspace}
\usepackage[inline]{enumitem}
\usepackage[nohyperlinks,nolist]{acronym}
\usepackage{todonotes}
\usepackage{amsmath}
\usepackage{amssymb}
\usepackage{booktabs}
\usepackage{multirow}
\usepackage{makecell}
\usepackage{subfig}
\usepackage{multicol}

\usepackage{hyperref}

\makeatletter
\DeclareRobustCommand\onedot{\futurelet\@let@token\@onedot}
\def\@onedot{\ifx\@let@token.\else.\null\fi\xspace}
\def\eg{e.g\onedot} 
\def\ie{i.e\onedot}

\def\etal{et al\onedot}

\newcommand{\styleganII}{StyleGAN2\xspace}
\newcommand{\stylegan}{StyleGAN\xspace}

\newcommand{\docufcn}{Doc-UFCN\xspace}
\newcommand{\emanet}{EMANet\xspace}
\newcommand{\transunet}{TransUNet\xspace}
\newcommand{\datasetgan}{DatasetGAN\xspace}
\newcommand{\miou}{mIoU\xspace}

\begin{document}

\title{Synthesis in Style: Semantic Segmentation of Historical Documents using Synthetic Data}

\begin{acronym}
  \acro{OCR}{Optical Character Recognition}
  \acro{GAN}{Generative Adversarial Network}
  \acro{IoU}{Intersection over Union}
  \acro{WPI}{Wildenstein Plattner Institute}
\end{acronym}

\author{
\IEEEauthorblockN{
  Christian Bartz\textsuperscript{\textasteriskcentered} Hendrik Raetz\textsuperscript{\textasteriskcentered}, Jona Otholt, Christoph Meinel, Haojin Yang}
\IEEEauthorblockA{
  Hasso Plattner Institute, University of Potsdam \\
  Potsdam, Germany \\
  \{firstname.lastname\}@hpi.de}
}

\maketitle
\begingroup\renewcommand\thefootnote{\textasteriskcentered}
\footnotetext{Equal contribution}
\endgroup

\begin{abstract}
    One of the most pressing problems in the automated analysis of historical documents is the availability of annotated training data.
    The problem is that labeling samples is a time-consuming task because it requires human expertise and thus, cannot be automated well.
    In this work, we propose a novel method to construct synthetic labeled datasets for historical documents where no annotations are available.
    We train a \stylegan model to synthesize document images that capture the core features of the original documents.
    While originally, the \stylegan architecture was not intended to produce labels, it indirectly learns the underlying semantics to generate realistic images.
    Using our approach, we can extract the semantic information from the intermediate feature maps and use it to generate ground truth labels.
    To investigate if our synthetic dataset can be used to segment the text in historical documents, we use it to train multiple supervised segmentation models and evaluate their performance.
    We also train these models on another dataset created by a state-of-the-art synthesis approach to show that the models trained on our dataset achieve better results while requiring even less human annotation effort.
\end{abstract}

\IEEEpeerreviewmaketitle

\section{Introduction}
\label{sec:introduction}

\noindent

For the majority of history, humanity gathered its information in analog form and stored them in archives.
With the emergence of digital methods, more and more of these archives digitize their documents to preserve them for generations to come.
An additional benefit of this digitization is better indexing, which can help historians in their research.
However, the large amount of digitized documents (sometimes millions of scanned pages per archive) makes manual analysis impractical and calls for automation.

With the introduction of deep learning into the area of document analysis~\cite{oliveira_dhsegment_2018,boillet_multiple_2020,bartz_synthetic_2020,moysset_full-page_2017}, it is becoming more and more feasible to analyze large quantities of documents of digitized archives effectively.
In this paper, we focus on semantic segmentation of documents and try to extract three classes: background, printed text, and handwritten text.
Extracting these classes from a document is useful for two reasons:
It may help archivists identify pages where potentially valuable, handwritten annotations are located.
Additionally, it serves as a helpful preprocessing step before applying character recognition algorithms.

Many existing state-of-the-art approaches train machine learning models with annotated real-world data (more on related work in \autoref{sec:related_work}).
However, annotating real-world data is a highly time-consuming and thus cost-intensive task.
Therefore, adapting these methods to a new unlabeled dataset becomes inefficient due to the high expenses for gathering annotations.
The goal of our work is to perform semantic segmentation on document images without the need to label large amounts of images manually.

Synthetic data can be a way to circumvent this requirement.
However, while synthetic image generation is a well-researched topic, generating the corresponding labels is a problem that is far from solved.
In some subdomains, such as scene text detection, synthetic data has already been in use for years~\cite{gupta_synthetic_2016,long2020unrealtext,wang2011end}.
Still, the generated samples look artificial, and these approaches do not generalize to other applications.
\datasetgan~\cite{dataset_gan}, a more generalized approach relying on \acp{GAN}, creates realistic samples but requires a relatively high amount of manual labeling at pixel level.
To further reduce the manual labeling effort, we propose a method that only requires high-level annotations, reducing the annotation time by eight times.
In addition, we show that synthetic data generated with our approach is better suited for training semantic segmentation models.
On the compiled datasets, our method outperforms \datasetgan on all three segmentation models that were evaluated by us.
Similarly to \datasetgan, our approach is not limited to a specific domain but can be applied to every field where enough data is available to train a functioning \stylegan network.

Our proposed pipeline consists of several steps:
First, we directly use scanned documents to train a \ac{GAN} to synthesize artificial documents that look as realistic as possible.
Additionally, we use the knowledge of the trained model for generating a label image that contains the semantic class for each pixel of the synthesized image.
Specifically, we utilize the generative capabilities of a \stylegan model~\cite{karras_style-based_2019,karras_analyzing_2020} and the observation that intermediate layers of a \stylegan model might encode the semantic class of pixels~\cite{collins_editing_2020}.
By applying an unsupervised clustering algorithm to these intermediate layers, we can create clusters that represent the semantic classes of the corresponding pixels.
At this point, human intervention is required to classify these clusters.
However, using the interface provided by us, this task does not take more than \num{30} minutes per trained \stylegan model.
Thus, we can use our system to synthesize a large, annotated dataset, which we use to train an additional semantic segmentation model.
A more detailed description of our proposed pipeline can be found in \autoref{sec:method}.

In \autoref{sec:experiments}, we analyze the capabilities and weaknesses of our synthesis approach.
For this, we create an artificial dataset that we use for the training of multiple recently-proposed segmentation models~\cite{boillet_multiple_2020,ema_net,chen2021transunet}.

In \autoref{sec:conclusion}, we conclude the findings of this paper and provide a short intro into further extensions of our proposed idea.

In summary, the contributions of this paper are as follows:
\begin{enumerate*}[label={\arabic*)}]
    \item We propose a novel approach for the semi-automatic synthesis of training data for the semantic analysis of documents based on the intermediate features of generative models and show that it can be used to create fine-grained semantic label images.
    \item For evaluation purposes, we provide a labeled dataset of high-resolution historical documents that includes detailed annotations.
    \item In our experiments, we train multiple segmentation models entirely on artificial datasets and show that they can successfully segment real-world documents, outperforming previous work.
    \item We provide our code and dataset to the community for further experimentation\footnote{\url{https://github.com/hendraet/synthesis-in-style}}.
\end{enumerate*}

\section{Related Work}
\label{sec:related_work}

The field of semantic structure analysis of historical documents features a wide variety of research topics, \eg, page detection, page segmentation, layout analysis and line segmentation~\cite{ oliveira_dhsegment_2018,boillet_multiple_2020,stewart_document_2017}.
However, these methods rely on annotated training data that is tailored to their specific use case.
Thus, it is not possible to easily apply them to datasets where no suitable annotations are available because acquiring labeled data is usually a very costly process that requires a lot of human work.

Looking at other fields, we can see that using synthetic data could be the key to solve the problem at hand.
Synthetic data has been successfully employed in the field of scene text detection~\cite{gupta_synthetic_2016} and scene text recognition~\cite{jaderberg_reading_2016}, where artificial text was rendered on different natural backgrounds in plausible positions.
Recent work~\cite{bartz_synthetic_2020} also shows that artificial data can be utilized in the area of historical document analysis.

The idea to extract semantic information from the intermediate feature maps of a \stylegan network was also shown in a concurrent work by Zhang~\etal~\cite{dataset_gan}.
This method is denoted \emph{DatasetGAN} and can be used to synthesize labels for natural images, such as faces, cars, and animals.
To simultaneously generate labels and artificial images, first, \stylegan's latent codes are recorded while creating a set of samples.
In a second step, they let a human annotator manually generate labels for these samples, which they use to train an ensemble of small classification networks to predict label images based on latent codes.
Subsequently, when using \stylegan to generate new samples, the trained ensemble automatically creates labels based on the latent codes.

A similar approach was later introduced by Li~\etal~\cite{semantic_gan} and successfully used for the segmentation of medical images (CT/MRT scans) and images of human faces.
They add another branch to a \styleganII network that is trained to output the label image together with a synthesized image.
Additionally, an encoder network is trained, which embeds a target image as noise that can be used by the \stylegan network to reproduce the original image.
Thus, the label image for the encoded image is produced as well.

Both approaches significantly reduce the time needed to create labeled datasets.
However, human intervention is still required to annotate images.
Depending on the complexity of the problem, this might still take several hours and might be challenging for inexperienced annotators.
Additionally, the examined natural images have different properties than scanned documents.
Images of text feature specific semantics because they contain important details, such as fine strokes or punctuation.
Thus, it is worth investigating if \stylegan can learn these characteristics so that it can produce artificial datasets for historical document analysis.

\section{Method}
\label{sec:method}

\noindent
In contrast to recent document analysis methods, we focus on training a segmentation model entirely on synthetic data.
This data is generated by a generative model that is trained to closely replicate the real distribution.
Using synthetic data allows us to use well-established supervised learning methods for the semantic segmentation of document images.
The small domain gap between the synthetic and real data allows us to apply our trained models directly to our target data distribution.
In summary, our method leverages synthetic data to obtain a segmentation model that is tailored to the real data, even though no matching annotations are available for it.
In this section, we introduce our \stylegan-based data synthesis pipeline.
Furthermore, we introduce the semantic segmentation networks that we use in our experiments.

\subsection{Data Synthesis Pipeline}
\label{subsec:data_synthesis_pipeline}

Our proposed data synthesis pipeline consists of three basic steps.
In the first step, we train a \stylegan model on the samples of the database we want to analyze.
We chose to use \stylegan because it reaches state-of-the-art results in unconditional image generation and exhibits many interesting properties that allow us to control the image generation, as shown in several prior works~\cite{collins_editing_2020,abdal_image2stylegan_2019,abdal_image2stylegan_plus_2019,pidhorskyi_adversarial_2020,nie_semi-supervised_2020,bartz_one_2020}.

\begin{figure*}[t]
    \centering
    \includegraphics[width=0.9\textwidth]{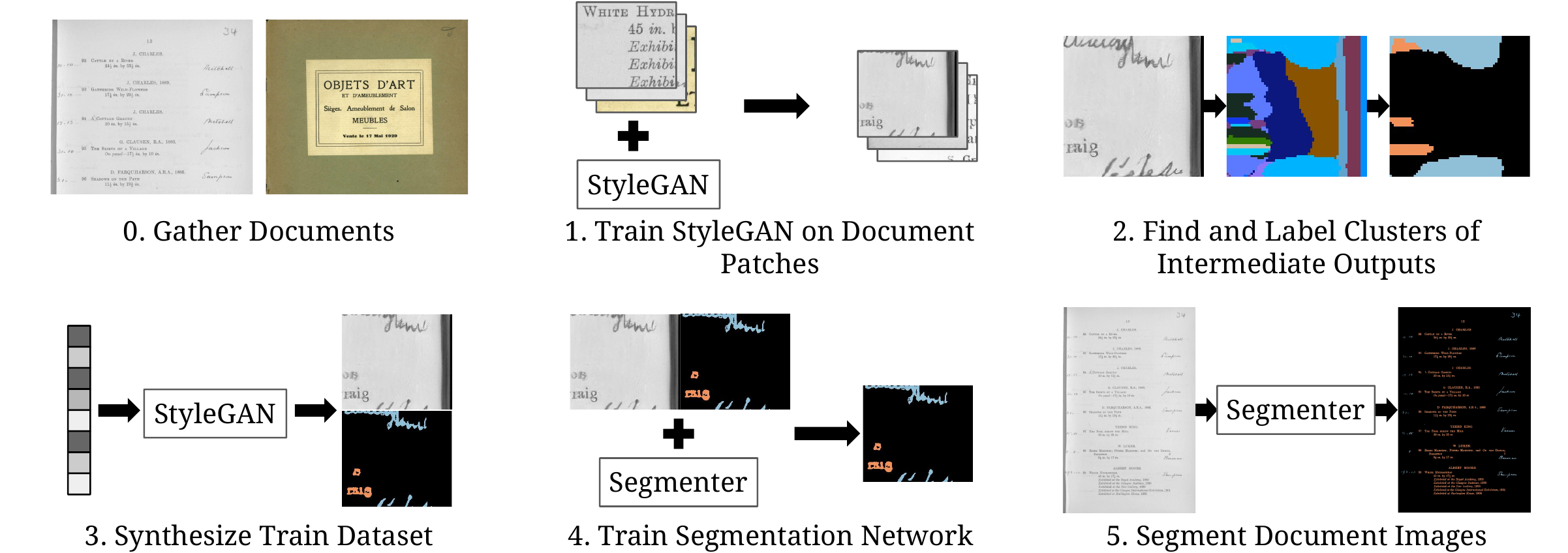}
    \caption{
        Depiction of our proposed pipeline.
        Our pipeline consists of the following steps:
        0. We gather scans of documents.
        1. We train a \stylegan model to generate document patches that look as similar, as possible, to real patches extracted from our document corpus.
        2. We use an unsupervised clustering algorithm on the intermediate outputs of the synthesis network of our trained \stylegan model and annotate the found clusters manually
        3. We use the \stylegan model from step 2 and the classified clusters to synthesize a training dataset.
        4. We use the synthesized training data to train an off-the-shelf segmentation network on patches of documents.
        5. We apply the trained segmentation network on the real document images
    }
    \label{fig:proposed_pipeline}
\end{figure*}

In our second step, we make use of the observation that intermediate layers of \stylegan can encode semantic information about the class of each pixel in the resulting image~\cite{collins_editing_2020}.
Based on this idea, we create an algorithm that uses the output of these intermediate layers for the synthesis of label images.
In the last step, we use our trained \stylegan model and our analysis algorithm to synthesize a large-scale, fully annotated dataset for document segmentation that we then use for the training of an off-the-shelf semantic segmentation network.
In the following, we explain our pipeline using the example of segmenting and classifying printed and handwritten text in document images.
In~\autoref{fig:proposed_pipeline}, we show how data synthesis is utilized in our approach.

\subsubsection{Training of \stylegan}
\label{subsubsec:training_of_stylegan}

First, we train a \ac{GAN} based on the \stylegan architecture~\cite{karras_style-based_2019} using the original document images.
The \stylegan architecture proposed by Karras~\etal consists of three main components.
The core component of the model is the synthesis network, which uses progressive growing~\cite{karras_progressive_2018}, enabling it to synthesize high-resolution images of high quality.
It receives style guidance from the mapping network that maps a latent vector $z \in \mathcal{Z}$ with $\mathcal{Z} \in \mathbb{R}^n$ into an intermediate latent space $r \in \mathcal{R}$ with $\mathcal{R} \in \mathbb{R}^n$.
These vectors are then fed into the synthesis network, allowing it to generate diverse images.
The last component is another input to the synthesis network called \emph{stochastic noise}, which helps to generate stochastic details.
In the case of face generation, these details can be freckles or hair, whereas, in the case of document generation, this noise can, \eg, influence the design of characters.
We use \styleganII~\cite{karras_analyzing_2020}, an improved version of \stylegan that addresses several weaknesses of the original model.
For further information about \stylegan and \styleganII, please refer to~\cite{karras_style-based_2019} and~\cite{karras_analyzing_2020}.

We do not use entire document images as input to \stylegan.
Instead, we divide each document image into multiple patches and train our \stylegan model to synthesize patches of images.
We choose to synthesize patches because it is simpler to create a realistic-looking patch than to generate a full document that resembles real ones.
Using patches allows us to analyze documents at a high-resolution without compromising the granularity of the results.
Additionally, this approach allows us to work on documents of varying sizes without having to account for different aspect ratios.
Furthermore, documents combine various aspects, e.g., areas with printed or handwritten text, images, text decorations, and scanning margins.
Patches allow the generator to concentrate on specific ones because it does not have to include many variations or combinations of these aspects simultaneously.
However, using patches adds extra computational work during inference and might also cause inconsistencies when assembling the patches after performing semantic segmentation with a segmentation network.

\subsubsection{Analysis of the Trained \stylegan Model}
\label{subsubsec:analysis_of_trained_stylegan_model}

Once we have trained the \stylegan model, we use it to synthesize annotated training data for training a semantic segmentation network.
\stylegan was designed to synthesize realistic RGB images, \ie, similar to the provided input data.
Thus, at first glance, \stylegan seems unsuitable to synthesize realistic-looking samples and the corresponding label images. % Maybe put the paragraph in sec. 4 l.83 here?
However, we can deduce the label information from the feature maps within the internal layers of \stylegan.
During the generation of a sample, \stylegan encodes the semantic class of pixels in the intermediate layers of the synthesis network.
This behavior was first described by Collins~\etal~\cite{collins_editing_2020}, where they used this property to perform semantically meaningful local edits of faces.
They found that if an unsupervised clustering algorithm, such as spherical k-Means clustering~\cite{buchta_spherical_2012}, is applied to the activations of each \stylegan block in the synthesis network, semantic classes of each pixel can be determined.

\begin{figure}[t]
    \centering
    \includegraphics[width=\linewidth]{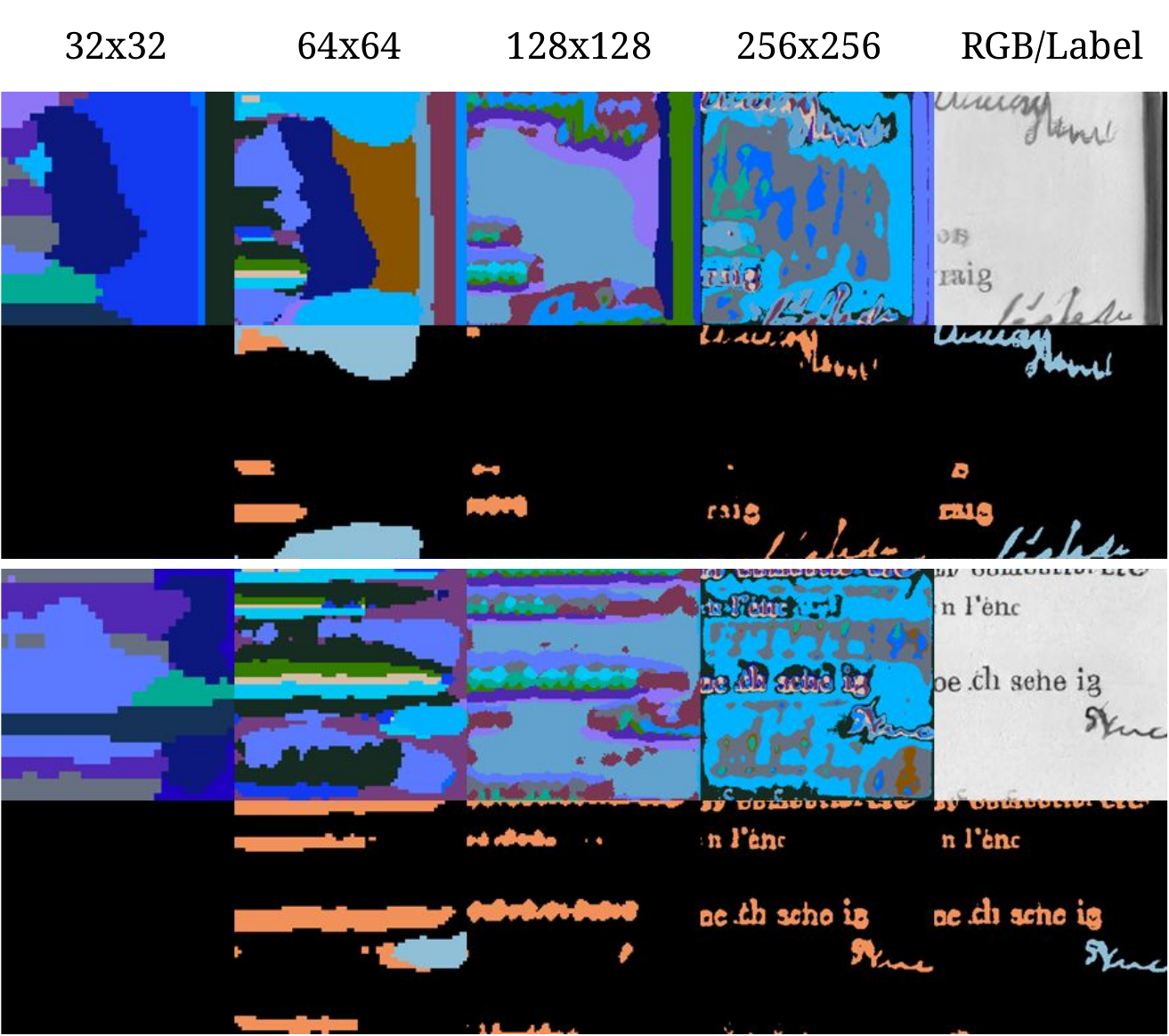}
    \caption{
        Visualization of the clustered intermediate layers and their corresponding labels for two synthesized samples (best viewed digitally and in color).
        The first four columns show the output of spherical k-means (20 clusters) on the intermediate layers of the synthesis network.
        The top row for each sample shows the original clusters, and the labeled counterparts are displayed directly below.
        The rightmost images depict the final image (top) and the matching label, which was created by combining the intermediate labels (bottom).
        In the label images, the blue color represents handwritten text, the orange color corresponds to printed text, and black stands for background.
        Since the last intermediate layer only contains structural information, the orange color only indicates the location of text and not the textual class.
        In practice, we have two layers for each size and layers that are smaller than $32 \times 32$.
        (These were omitted due to space limitations.)
    }
    \label{fig:clusters_and_labels}
\end{figure}

We follow the approach of Collins~\etal and examine the clusters of a \stylegan model trained on patches of document images that contain handwritten and printed text (Step 2 in~\autoref{fig:proposed_pipeline}).
We provide some samples and the results of clustering with spherical k-Means in~\autoref{fig:clusters_and_labels}.
The provided samples clearly show that certain intermediate layers of \stylegan's synthesis network encode the semantic class of printed or handwritten text very well.
However, these semantic clusters cannot be used directly for the synthesis of label images, because multiple clusters belong to a single class.
Also, if we were to use only one of these intermediate images for deciding the semantic class of each pixel, we would encounter several problems:
\begin{enumerate*}[label={(\arabic*)}]
	\item If we were to use an output of an early layer in the synthesis network, the resolution of our identified text regions in the resulting label image could be very low.
	\item The classes found within one layer of the synthesis network are not always accurate, thus, we have to rely on the output of multiple layers of the synthesis network.
\end{enumerate*}

To remedy these problems, we require a human annotator to examine the clusters found by spherical k-Means.
First, the last layers of the network (size $256 \times 256$) have to be analyzed by the annotator.
In this step, the main task is to separate the text clusters from background clusters to extract detailed structural information about the synthesized text.
However, at this stage of the network, the model only focuses on the texture and not on the shape~\cite{geirhos_imagenet-trained_2018}, thus distinction between handwritten and printed text is not possible.
Therefore, the second task of the annotator is to identify the feature maps that contain semantic information (usually layers of size $64 \times 64$ or $128 \times 128$) and determine the corresponding class for each of the detected clusters.
In very low-resolution feature maps, the clusters do not correspond to text regions, so they are not useful for creating label images.
\autoref{fig:clusters_and_labels} shows how the final annotated intermediate feature maps might look like.

We provide an algorithm that uses the annotations to combine the semantic information of the lower layers with the structural information of the last layers to create a label image of sufficient detail.
It might be necessary to adapt this algorithm for usage in different scenarios depending on the corresponding \stylegan output.

We found that examining a maximum of \num{100} images suffices to accurately determine the classes of clusters.
To speed up the labeling process and keep the annotation time minimal, we provide a simple labeling tool in the code repository belonging to this publication.

\subsubsection{Synthesis of a Large Scale Dataset}
\label{subsubsec:synthesis_of_large_scale_dataset}

Once the clusters are labeled, we can run the \stylegan model in combination with our algorithm to create a large-scale training dataset.
Here, we draw vectors from a uniform distribution and pass them to our \stylegan model to produce synthetic document patches and label images.
The resulting dataset might be imbalanced because it could contain more images only depicting background information than images displaying text.
This happens because it is highly probable that most patches used for the training of our \stylegan model do not contain any text.
However, we can later balance the dataset by using the synthesized annotations.

\subsection{Semantic Segmentation Network}
\label{subsec:semantic_segmentation_network}

The second to last step of our pipeline is the training of semantic segmentation networks using our synthetic training data.
In our experiments, we decided to use three different segmentation networks.
The first is \docufcn, a semantic segmentation network that reaches state-of-the-art results in the line segmentation task~\cite{boillet_multiple_2020}.
The network by Boillet~\etal is based on a U-Net~\cite{ronneberger_u-net:_2015} architecture, uses custom dilated convolution blocks and has less trainable parameters than other state-of-the-art networks.
EMANet~\cite{ema_net}, the second network, was originally designed for the semantic segmentation of natural image datasets, such as PASCAL VOC~\cite{pascal_voc}.
The backbone of this model is a ResNet-101 that incorporates the eponymous \emph{EMAUnits}, which introduce a special attention mechanism.
As third network, we choose TransUNet~\cite{chen2021transunet}, which was developed for the segmentation of medical images.
As the name implies, it employs the U-Net architecture while using a Transformer~\cite{transformer} as an encoder.
We chose these networks because they reach state-of-the-art results in their respective field and exhibit different strengths that we want to evaluate.
\docufcn is specifically designed for segmentation tasks on historical documents, EMANet reaches state-of-the-art performance on widely researched benchmark datasets, and TransUNet can produce detailed segmentations required in the field of medical segmentation.

To enhance the performance of the segmentation networks, we assert that input images are grayscale before feeding them into the training pipeline.
We enhance the diversity of the training data by randomly augmenting the input images.
These augmentation operations include: cropping, shearing, shifting, slight distortion, rotation, contrast change and color inversion.

\section{Experiments}
\label{sec:experiments}

In principle, we can apply our proposed pipeline to any unlabeled dataset and create semantic segmentation models that are custom-fit to the dataset.
We show in our experiments that this applies to real-world images of auction catalogs that we extracted from an art-historical dataset.
In this section, we first introduce the dataset we use to evaluate our pipeline.
Afterwards, we provide a detailed description of our experimental setup.
We finish this section by presenting our results and discussing the possibilities and limitations of our proposed method.

\subsection{Benchmark Dataset}
\label{subsec:dataset}

\begin{figure*}[!t]
  \centering
  \includegraphics[width=0.95\linewidth]{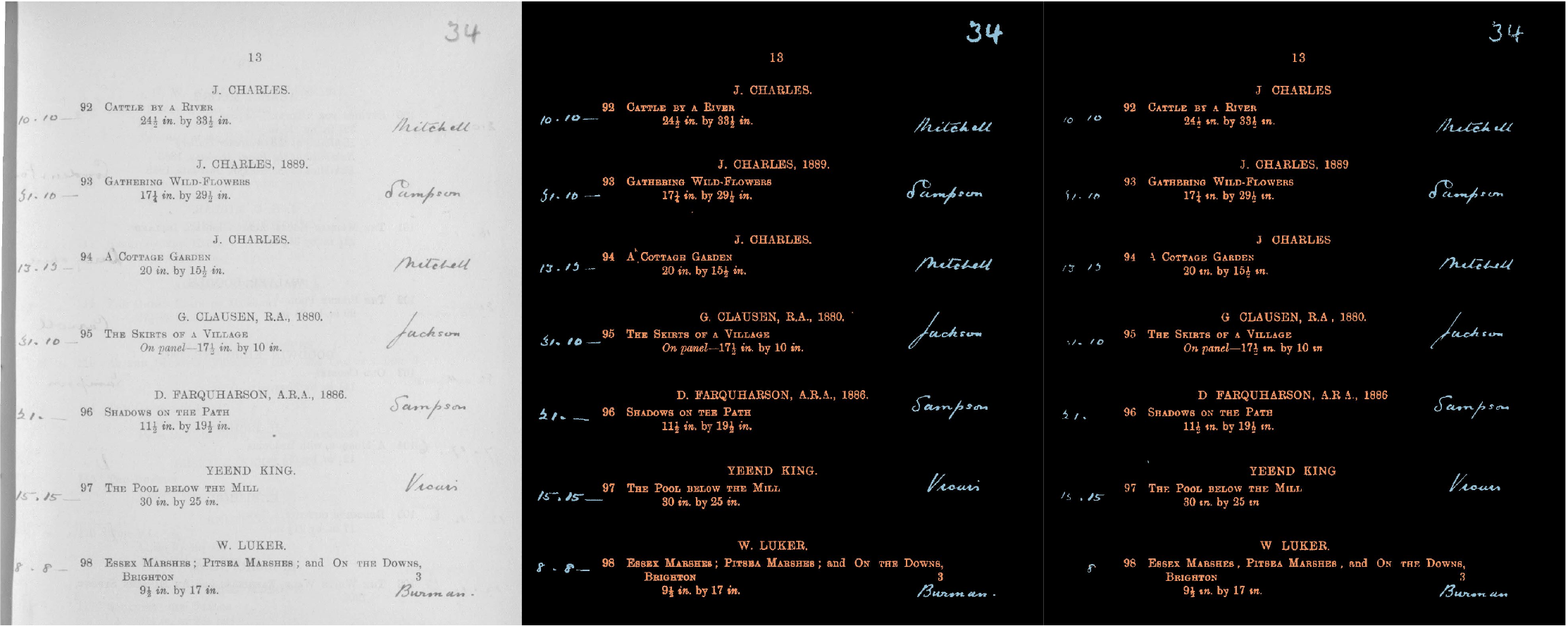}
  \caption{Image cutouts of a sample from the in-domain split of the evaluation dataset, the corresponding ground truth and our prediction made with \transunet on the right. More examples can be found in the supplementary material.}
  \label{fig:dataset_examples}
\end{figure*}

To the best of our knowledge, there are no datasets that include fine-grained, pixel-level annotations of historical documents.
Existing datasets were built for less detailed segmentation tasks, such as line segmentation or layout analysis~\cite{gruning_read-bad_2018, boillet_horae_2019, chron_seg}.
However, we want to evaluate if the trained segmentation models can generate even more precise segmentations.
Thus, we manually labeled 32 high-resolution images of scanned document pages obtained from the archive of the \ac{WPI}.
These documents are pages of auction catalogs that contain mainly printed text but also handwritten letters and annotations that are especially interesting for researchers.
In addition to the printed and handwritten text, which comes in various fonts, sizes, and orientations, the catalogs also contain images of paintings.
\autoref{fig:dataset_examples} shows an example image and the corresponding ground truth.
For better evaluation, we decided to split the dataset in two parts: in-domain and out-of-domain.
The images that are considered in-domain are representative for the majority of the samples in the dataset we used to train our generative models.
The out-of-domain images feature more diverse text and page layouts that may not be as well reflected by the learned distribution of the generative models.

\subsection{Experimental Setup}

We use our pipeline to synthesize \num{100000} training images with the corresponding ground truth.
Afterwards, this dataset is balanced so that it contains an equal amount of images containing handwriting and images containing printed text.

We perform post-processing on the segmented image patches to remove noise and improve the confidence of the predictions.
For this, we use two hyperparameters that are used by \docufcn: a threshold for keeping a text prediction (\emph{minimum confidence}) and a threshold for discarding small connected components of text (\emph{minimum contour area}).
Although these hyperparameters are also used during the training of \docufcn, we found that we can improve our results by applying them to all segmentation models during evaluation.
An additional hyperparameter is called \emph{patch overlap factor}, which steers the fragmentation of the original images.
If the overlap factor is larger than \num{0.0}, pixels will be classified multiple times because they are contained in different patches.
During the reassembling process, the most confident of these predictions determines the class of a pixel.
We find the optimal hyperparameters with an exhaustive grid search.
The resulting parameters we used for our experiments can be found in the appendix.

\subsection{Results and Discussion}

In our experiments, we want to evaluate the following:
\begin{enumerate*}[label={(\arabic*)}]
    \item which segmentation network is best suited for the segmentation of our benchmark dataset, and
    \item the performance of our segmentation method compared to \datasetgan~\cite{dataset_gan}.
\end{enumerate*}

\subsubsection{Performance of Segmentation Networks}

\begin{table}
	\centering
	\caption{Performance of the segmentation models on the in-domain split of the benchmark dataset.}
	\label{tab:in_domain}
	\begin{tabular}{ l c c c c c c c }
		\toprule
		\multirow{2}{*}[-3pt]{\makecell{Segmentation \\ Model}} & & \multicolumn{3}{c}{Printed Text} & \multicolumn{3}{c}{Handwritten Text}\\
		\cmidrule(lr){3-5} \cmidrule(lr){6-8}
		& mIoU & IoU & Prec. & Recall & IoU & Prec. & Recall \\
		\midrule
		\docufcn & 0.66 & 0.70 & 0.83 & 0.82 & 0.28 & 0.32 & \textbf{0.70} \\
		TransUNet & \textbf{0.72} & \textbf{0.71} & \textbf{0.84} & 0.82 & \textbf{0.46} & 0.67 & 0.60 \\
		EMANet & 0.66 & 0.52 & 0.56 & \textbf{0.87} & 0.46 & \textbf{0.71} & 0.57 \\
		\bottomrule
	\end{tabular}
\end{table}

\begin{table}
	\centering
	\caption{Performance of the segmentation models on the out-of-domain split of the benchmark dataset.}
	\label{tab:out_of_domain}
	\begin{tabular}{ l c c c c c c c }
		\toprule
		\multirow{2}{*}[-3pt]{\makecell{Segmentation \\ Model}} & & \multicolumn{3}{c}{Printed Text} & \multicolumn{3}{c}{Handwritten Text}\\
		\cmidrule(lr){3-5} \cmidrule(lr){6-8}
		& mIoU & IoU & Prec. & Recall & IoU & Prec. & Recall \\
		\midrule
		\docufcn & 0.49 & 0.37 & \textbf{0.87} & 0.40 & 0.11 & 0.15 & 0.25 \\
		TransUNet & 0.54 & 0.47 & \textbf{0.87} & 0.50 & \textbf{0.17} & \textbf{0.38} & 0.24 \\
		EMANet & \textbf{0.56} & \textbf{0.52} & 0.67 & \textbf{0.70} & \textbf{0.17} & 0.27 & \textbf{0.30} \\
		\bottomrule
	\end{tabular}
\end{table}

\autoref{tab:in_domain} shows the performance of the segmentation models on the in-domain split of our benchmark dataset.
Besides the widely used mean \acs{IoU} (\miou), we report the class-wise \ac{IoU}, precision, and recall for handwritten and printed text\footnote{
    We have decided to leave out the class scores for background because it does not provide any added value for the evaluation.
    However, detailed results for all experiments can be found in the supplementary material.
}.

Overall, the results are satisfying.
\transunet shows the best performance while the other models do not trail far behind.
However, \emanet and \docufcn achieve higher recall values for printed and handwritten text, respectively.
Taking a closer look at class-wise metrics reveals that it is much easier for the models to correctly identify printed text compared to handwritten text.
A reason for this significant difference might be the underlying dataset used for the training of \stylegan.
The original images contain a lot of printed text and only occasionally feature handwritten annotations.
Thus, \stylegan will produce more samples depicting printed text.
Balancing the synthetic training dataset so that there are as many samples containing handwritten text as samples showing printed text mitigates this issue to a certain degree.
Overall, the dataset still contains significantly more pixels belonging to the printed text class.
Thus, during training, the segmentation models learn to detect printed text more reliably.

The results on the out-of-domain split are shown in \autoref{tab:out_of_domain}.
As expected, the \ac{IoU} is lower across the board, with handwritten text showing a more pronounced drop than printed text.
This is mostly due to a lower precision of the handwritten text classifications compared to the in-domain data, whereas for printed text the precision is almost unchanged.
A possible explanation is that the out-of-domain samples contain more objects that are easily misclassified as handwritten text, for example drawings, paintings, or unusual fonts.

\subsubsection{Comparison to \datasetgan}

\begin{table}
	\centering
    \caption{Comparison between our data synthesis approach and DatasetGAN for different segmentation models.}
	\label{tab:vs_dataset_gan}
	\begin{tabular}{ l l c c }
		\toprule
		\multirow{2}{*}[-3pt]{\makecell{Segmentation \\ Model}} &\multirow{2}{*}[-3pt]{\makecell{Synthesis \\ Method}}  
	    & \multicolumn{2}{c}{mIoU}\\
		\cmidrule(lr){3-4}
	    & & In-Domain & Out-of-Domain \\
		\midrule
        \docufcn & DatsetGAN & 0.65 & 0.45 \\
        & Ours & 0.66 & 0.49 \\
        TransUNet & DatsetGAN & 0.61 & 0.48 \\
        & Ours & \textbf{0.72} & 0.54 \\
        EMANet &  DatsetGAN & 0.52 & 0.43 \\
        &  Ours &0.66 & \textbf{0.56} \\
		\bottomrule
	\end{tabular}
\end{table}

In \autoref{tab:vs_dataset_gan}, we evaluate the performance of our dataset synthesis approach.
For this, we trained the segmentation models on a dataset synthesized using the \datasetgan method~\cite{dataset_gan}.
When comparing the mIoUs, it can be seen that the models trained using our method achieve better results on both the in-domain and the out-of-domain samples.
We believe that this is the case because \datasetgan was developed for a completely different domain.
Natural images often depict larger coherent structures, whereas the text in document images is a set of detailed characters where minor variations can change the semantic class.
As a result, the synthesized data produced by \stylegan is noisier and contains more ambiguities, such as partially drawn letters that make it harder to establish a consistent segmentation.
Our approach, which labels more samples but with a straightforward, cluster-based labeling procedure, can achieve a more consistent segmentation with less labeling effort.
To put labeling time in perspective: annotating an adequate dataset required by \datasetgan (10 images) took us around four hours, whereas labeling using our method only takes about 30 minutes.
It is possible that using more labeled samples would further improve the performance of \datasetgan, but this improvement would probably be disproportionate to the required labeling effort.

\section{Conclusion and Future Work}
\label{sec:conclusion}

In this paper, we have proposed a novel approach for synthesizing large-scale training datasets, which is suited (but not limited) to the analysis of historical document images.
Our approach works directly on scans of documents without the need to annotate large amounts of individual images.
However, we still require human intervention for the labeling of clusters, for which we provide an easy-to-use interface to keep the annotation time minimal.
This enables us to train fully-supervised machine learning models on datasets that do not have any annotations available.

In our experiments, we have shown that state-of-the-art segmentation models that have been trained on our synthesized datasets can produce satisfactory segmentations.
In addition, our proposed method outperforms the similar DatasetGAN~\cite{dataset_gan} due to its easy and consistent labeling process, while requiring less labeling effort.
These results prove that our approach can help to make machine learning more accessible by avoiding the high labeling cost associated with traditional supervised methods.

In the future, we wish to improve our pipeline and use it to analyze different datasets where no annotations are available.
To improve our pipeline, we are especially interested in removing the need for human intervention, reducing the amount of computational resources necessary for data synthesis, and improving the generalizability of models created with our proposed system.

\bibliographystyle{IEEEtran}
\bibliography{IEEEabrv,biblio}

\clearpage
\onecolumn

\appendix

\subsection*{More Information on Experimental Setup}

\begin{multicols}{2}
    \noindent
    Additional information on the experimental setup:
    We performed our experiments on systems with a GPU that has at least \SI{11}{\giga\byte} of RAM (e.g., using a Geforce GTX 1080Ti).
    For the training of our \stylegan model, we follow the hyperparameters set by Karras~\etal~\cite{karras_analyzing_2020}, but we set the initial learning rate to \num{0.001}, the number of iterations to \num{100000} and we use cosine annealing~\cite{Loshchilov_cosine_annealing} for updating the learning rate during training.
    We set the image size of \stylegan to $256 \times 256$ pixels for all of our experiments.

    During the training of \docufcn for segmentation, we follow the hyperparameters of Boillet~\etal~\cite{boillet_multiple_2020} by setting the initial learning rate to \num{0.005}, the threshold for keeping a text prediction (\emph{minimum confidence}) to \num{0.7}, and we discard connected components of text with an area of less than \num{50} pixels (\emph{minimum contour area}).
    EMANet~\cite{ema_net} is also trained using the proposed hyperparameters, namely an initial learning rate of \num{0.009}.
    Weight decay coefficients are set to \num{0.9} and \num{0.0001}.
    The hyperparameter configuration for TransUNet~\cite{chen2021transunet} is also taken from the referenced paper: learning rate of \num{0.01}, momentum of \num{0.9} and weight decay of \num{0.0001}.
    For the training of all our segmentation models, we use a batch size of \num{8} and cosine annealing for updating our learning rate during training.
\end{multicols}

\begin{table*}[ht]
	\centering
	\caption{Performance of the on the in-domain split across all generation models and segmentation models}
	\label{tab:appendix_in_domain}
	\begin{tabular}{ l l c c c c c c c c c c }
		\toprule
		\multirow{2}{*}[-3pt]{\makecell{Synthesis \\ Method}} & \multirow{2}{*}[-3pt]{\makecell{Segmentation \\ Model}} & & \multicolumn{3}{c}{printed text} & \multicolumn{3}{c}{handwritten text} & \multicolumn{3}{c}{background}\\
		\cmidrule(lr){4-6} \cmidrule(lr){7-9} \cmidrule(lr){10-12}
		& & mIoU & IoU & Prec. & Recall & IoU & Prec. & Recall & IoU & Prec. & Recall \\
		\midrule
		DatasetGAN & \docufcn & 0.655 & 0.643 & 0.827 & 0.743 & 0.326 & 0.624 & 0.406 & 0.995 & 0.997 & \textbf{0.998} \\
		& TransUNet & 0.615 & 0.609 & 0.832 & 0.695 & 0.239 & 0.294 & 0.562 & 0.996 & 0.997 & \textbf{0.998} \\
		& EMANet & 0.521 & 0.307 & 0.707 & 0.351 & 0.265 & 0.409 & 0.429 & 0.991 & 0.993 &\textbf{0.998} \\
		Ours & \docufcn & 0.658 & 0.696 & 0.825 & 0.817 & 0.281 & 0.320 & \textbf{0.700} & 0.996 & 0.998 & \textbf{0.998} \\
		& TransUNet & \textbf{0.725} & \textbf{0.714} & \textbf{0.843} & 0.824 & \textbf{0.463} & \textbf{0.672} & 0.598 & \textbf{0.997} & 0.998 & \textbf{0.998}\\
		& EMANet & 0.656 & 0.518 & 0.560 & \textbf{0.875} & 0.460 & 0.707 & 0.568 & 0.991 & \textbf{0.999} & 0.993\\
		\bottomrule
	\end{tabular}
\end{table*}

\begin{table*}[ht]
	\centering
	\caption{Performance of the on the out-of-domain split across all generation models and segmentation models}
	\label{tab:appendix_out_of_domain}
	\begin{tabular}{ l l c c c c c c c c c c }
		\toprule
		\multirow{2}{*}[-3pt]{\makecell{Synthesis \\ Method}} & \multirow{2}{*}[-3pt]{\makecell{Segmentation \\ Model}} & & \multicolumn{3}{c}{printed text} & \multicolumn{3}{c}{handwritten text} & \multicolumn{3}{c}{background}\\
		\cmidrule(lr){4-6} \cmidrule(lr){7-9} \cmidrule(lr){10-12}
		& & mIoU & IoU & Prec. & Recall & IoU & Prec. & Recall & IoU & Prec. & Recall \\
		\midrule
		DatasetGAN & \docufcn & 0.455 & 0.325 & 0.802 & 0.353 & 0.059 & 0.261 & 0.070 & 0.980 & 0.982 & 0.998 \\
		& TransUNet & 0.482 & 0.336 & 0.863 & 0.355 & 0.125 & 0.224 & 0.220 & 0.984 & 0.985 & 0.998\\
		& EMANet & 0.431 & 0.241 & 0.844 & 0.252 & 0.072 & 0.195 & 0.101 & 0.979 & 0.981 & \textbf{0.999}\\
		Ours & \docufcn & 0.488 & 0.375 & \textbf{0.875} & 0.396 & 0.106 & 0.154 & 0.255 & 0.983 & 0.987 & 0.996 \\
		& TransUNet & 0.541 & 0.466 & 0.866 & 0.502 & \textbf{0.173} & \textbf{0.384} & 0.240 & \textbf{0.985} & 0.987 & 0.998\\
		& EMANet & \textbf{0.557} & \textbf{0.523} & 0.675 & \textbf{0.699} & 0.167 & 0.270 & \textbf{0.303} & 0.983 & \textbf{0.992} & 0.991\\
		\bottomrule
	\end{tabular}
\end{table*}

\begin{table*}[ht]
	\centering
	\caption{Post-processing hyperparameters used for evaluation. The grid search was performed for the patch\_overlap\_factor \num{0.0} and \num{0.5}, the min\_confidence values \num{0.3}, \num{0.7}, \num{0.9} and the min\_contour\_area values \num{15}, \num{30}, and \num{55}}
	\label{tab:appendix_hyperparams}
	\begin{tabular}{ l l c c c }
		\toprule
		Synthesis Method & Segmentation Model & Minimum Confidence & Minimum Contour Area & Patch Overlap Factor\\
		\midrule
		DatasetGAN & \docufcn & 0.3 & 15 & 0.0 \\
		& TransUNet & 0.7 & 55 & 0.5 \\
		& EMANet & 0.3 & 15 & 0.5 \\
		Ours & \docufcn & 0.3 & 15 & 0.5 \\
		& TransUNet & 0.9 & 15 & 0.5 \\
		& EMANet & 0.3 & 15 & 0.5 \\
		\bottomrule
	\end{tabular}
\end{table*}

\begin{figure*}[!ht]
    \centering
    \subfloat{\includegraphics[width=\textwidth, height=0.3\textheight, keepaspectratio]{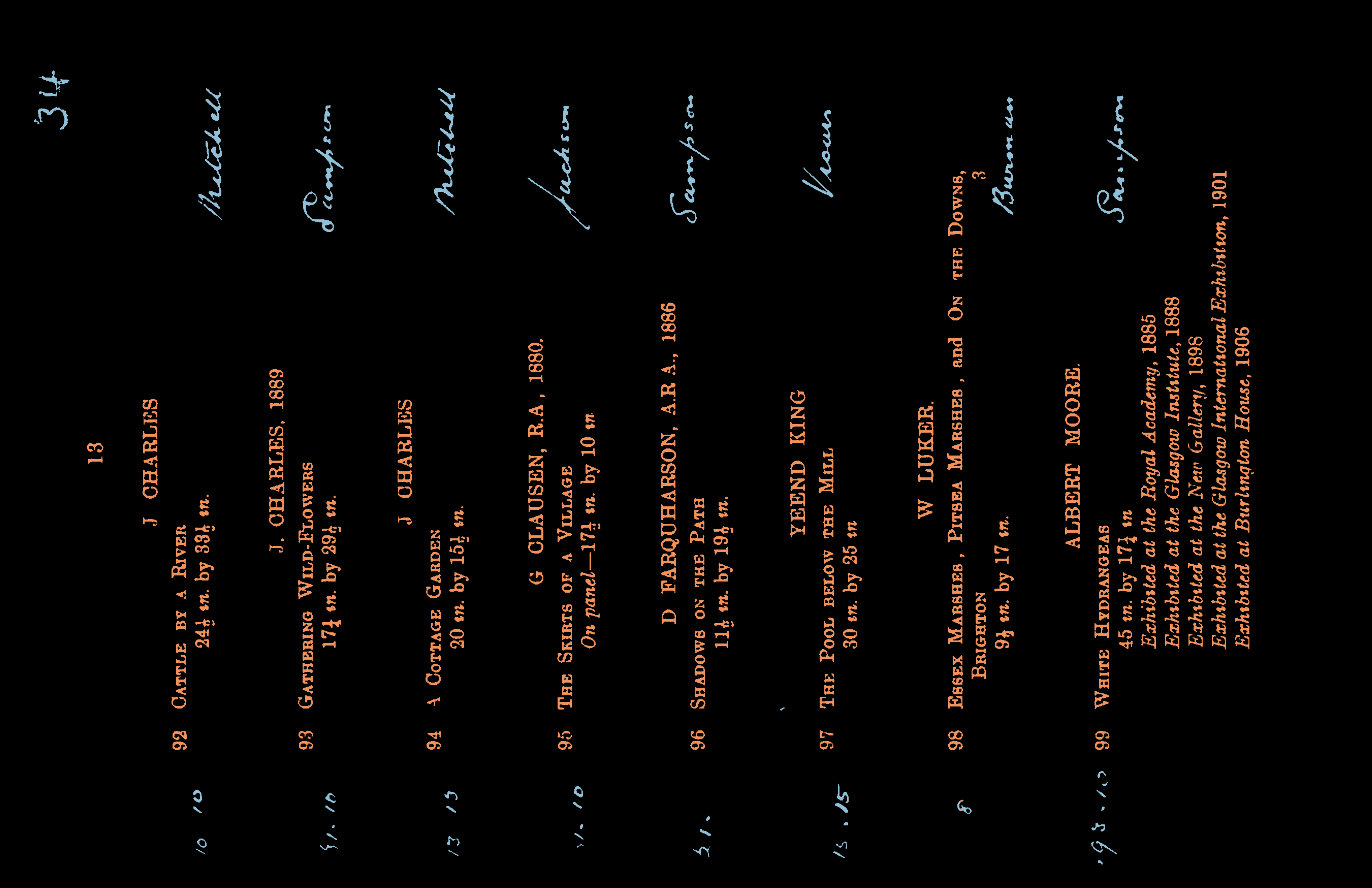}} \\
    \subfloat{\includegraphics[width=\textwidth, height=0.3\textheight, keepaspectratio]{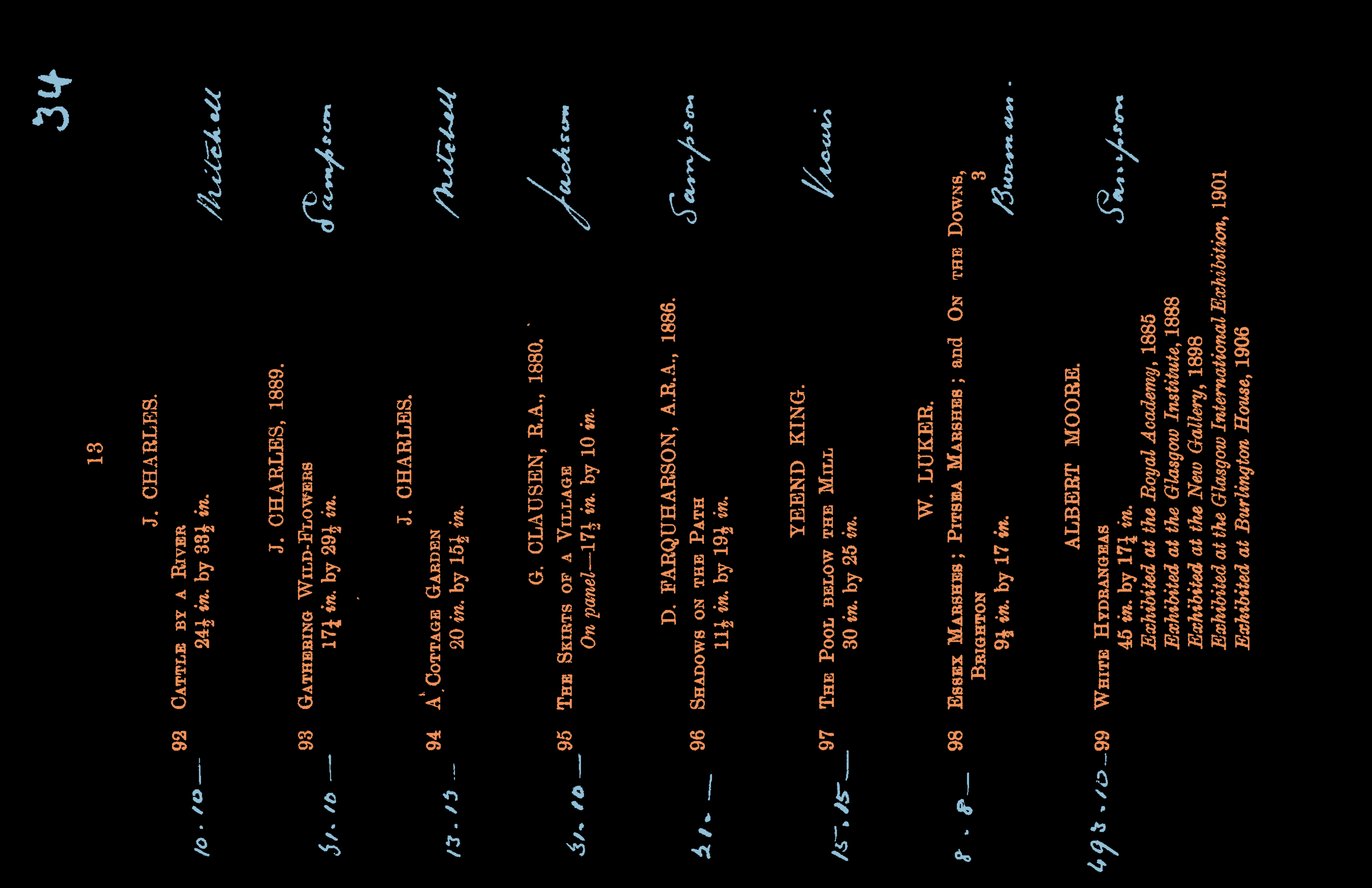}} \\
    \subfloat{\includegraphics[width=\textwidth, height=0.3\textheight, keepaspectratio]{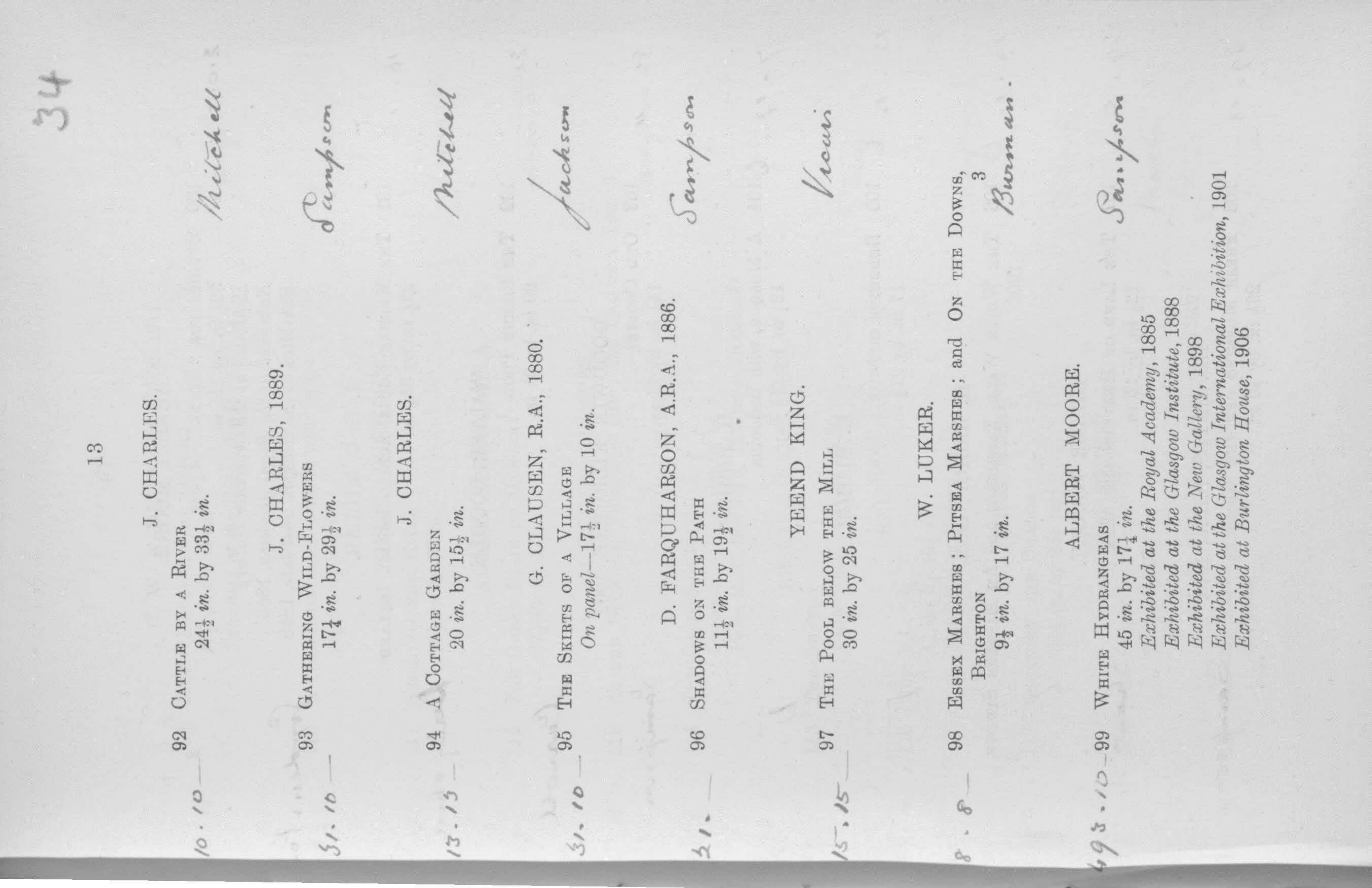}}
    \caption{Demonstration of our approach applied to an in-domain sample. Bottom: RGB image, Middle: Ground truth segmentation, Top: Prediction of TransUNet trained on synthetic data generated with our approach. The prediction is done on image patches that are then reassembled into the segmentation of the full image}
    \label{fig:qualitative_1}
  \end{figure*}

\begin{figure*}[!ht]
    \centering
    \subfloat{\includegraphics[width=\textwidth, height=0.3\textheight, keepaspectratio]{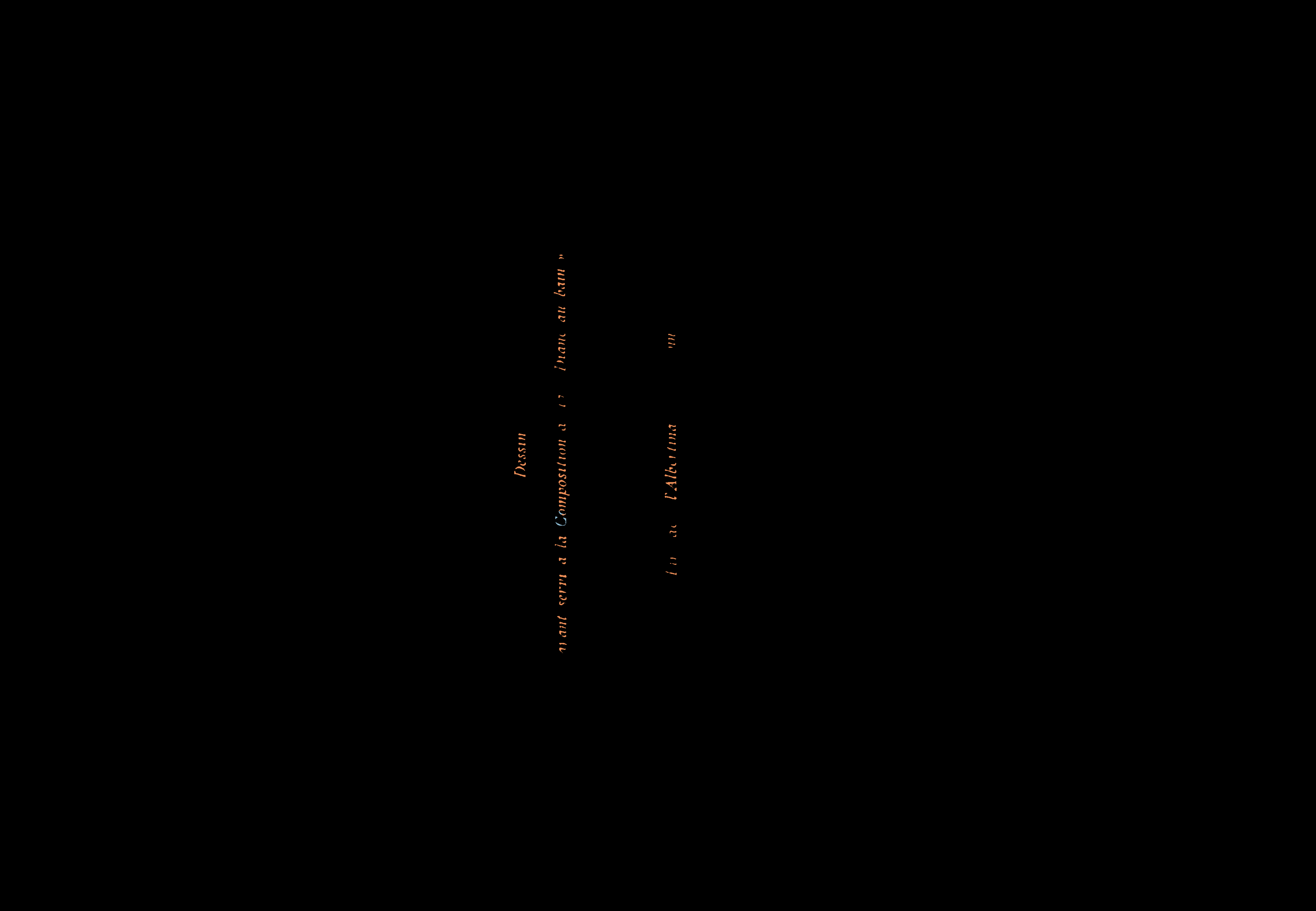}} \\
    \subfloat{\includegraphics[width=\textwidth, height=0.3\textheight, keepaspectratio]{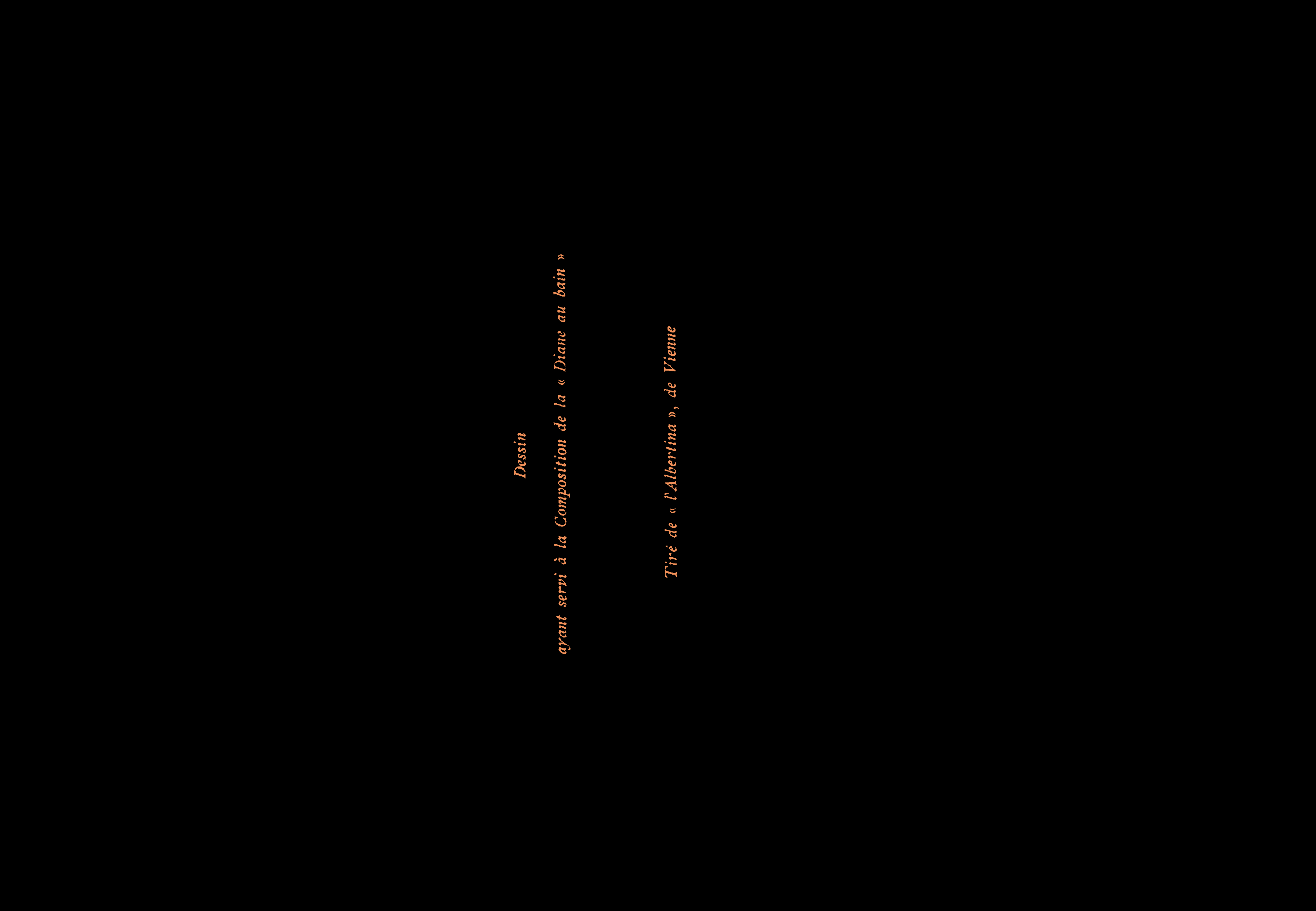}} \\
    \subfloat{\includegraphics[width=\textwidth, height=0.3\textheight, keepaspectratio]{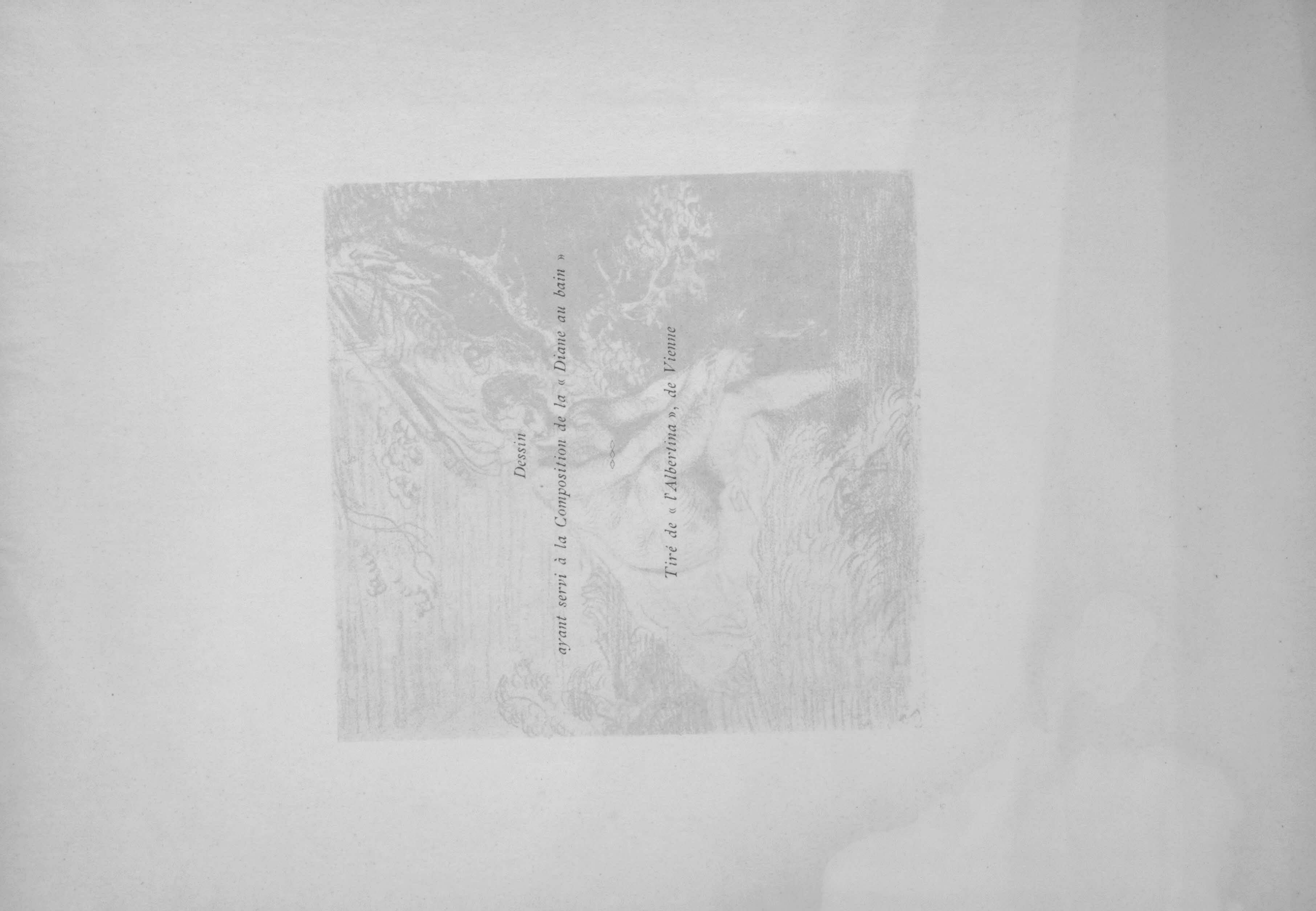}}
    \caption{Demonstration of our approach applied to an in-domain sample. Bottom: RGB image, Middle: Ground truth segmentation, Top: Prediction of TransUNet trained on synthetic data generated with our approach. The prediction is done on image patches that are then reassembled into the segmentation of the full image}
    \label{fig:qualitative_2}
  \end{figure*}

\begin{figure*}[!ht]
    \centering
    \subfloat{\includegraphics[width=\textwidth, height=0.3\textheight, keepaspectratio]{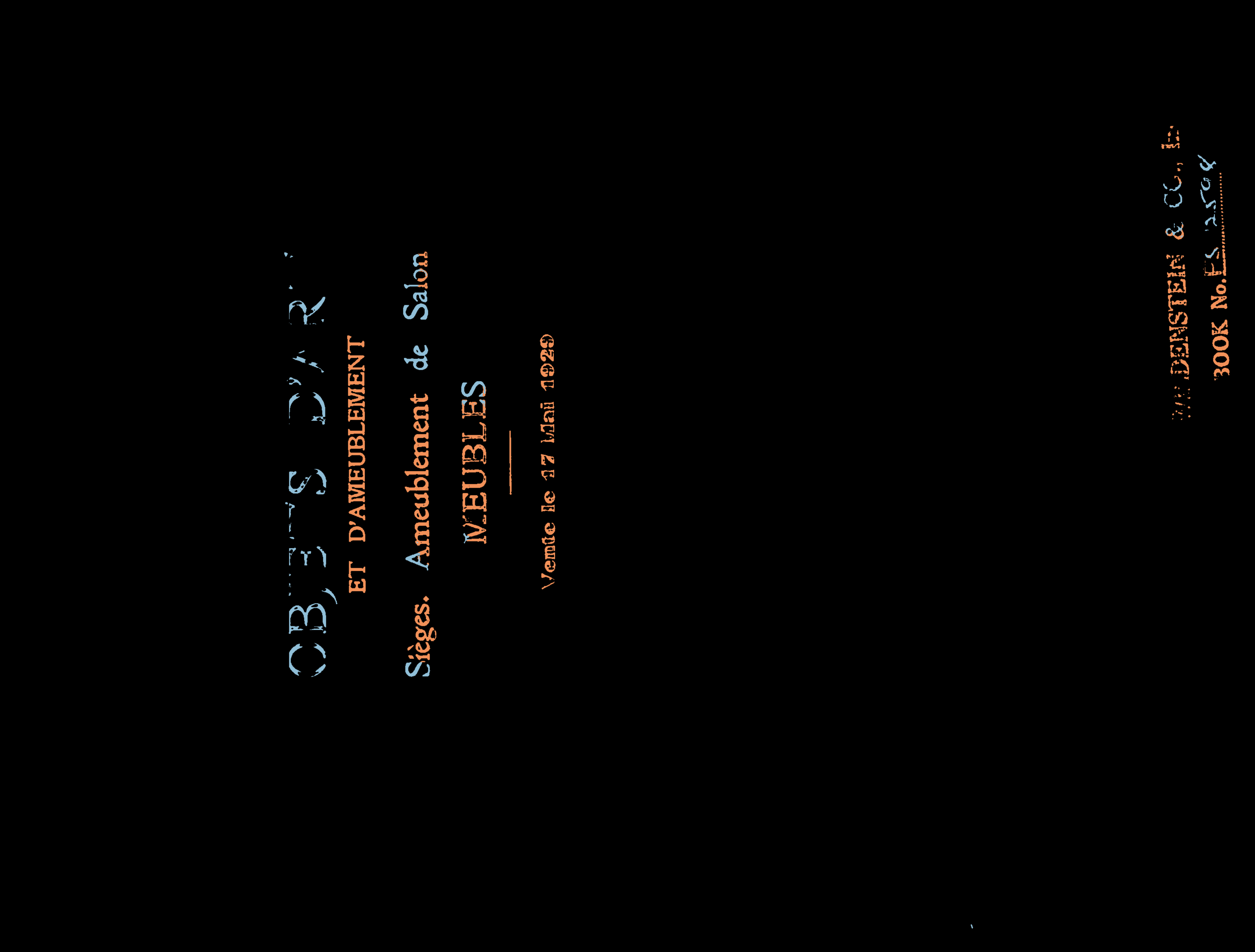}} \\
    \subfloat{\includegraphics[width=\textwidth, height=0.3\textheight, keepaspectratio]{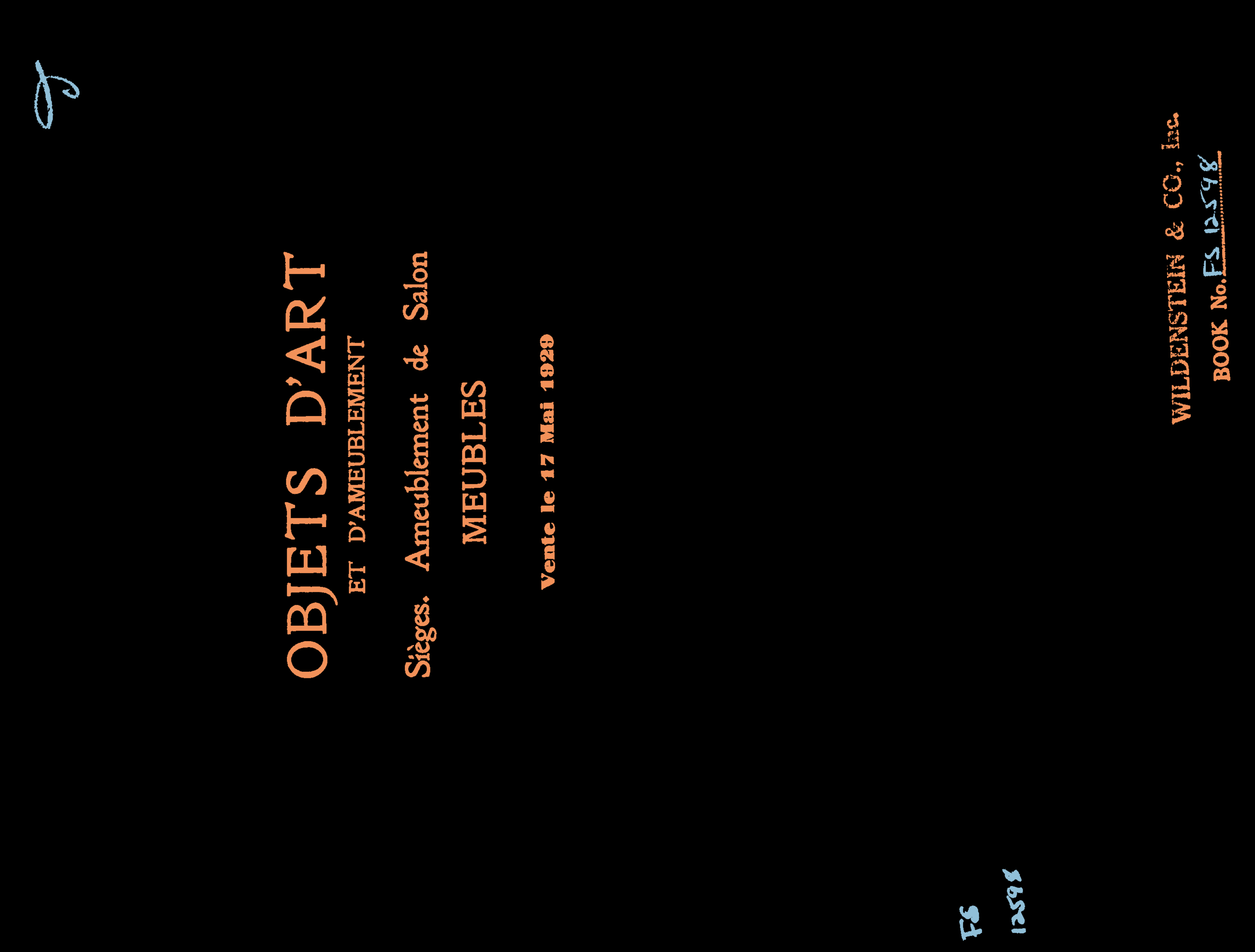}} \\
    \subfloat{\includegraphics[width=\textwidth, height=0.3\textheight, keepaspectratio]{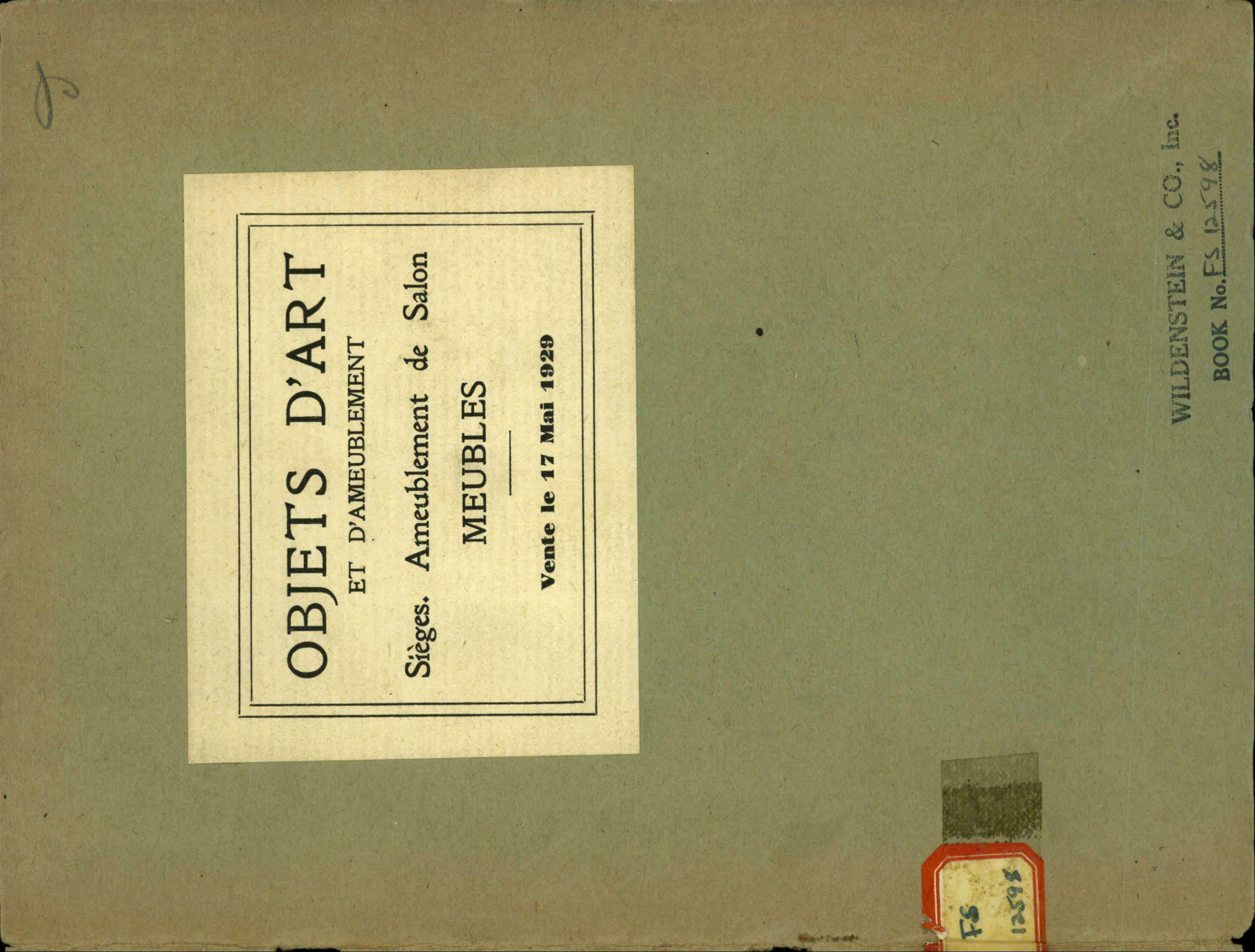}}
    \caption{Demonstration of our approach applied to an out-of-domain sample. Bottom: RGB image, Middle: Ground truth segmentation, Top: Prediction of TransUNet trained on synthetic data generated with our approach. The prediction is done on image patches that are then reassembled into the segmentation of the full image}
    \label{fig:qualitative_3}
  \end{figure*}

\begin{figure*}[!ht]
    \centering
    \subfloat{\includegraphics[width=\textwidth, height=0.3\textheight, keepaspectratio]{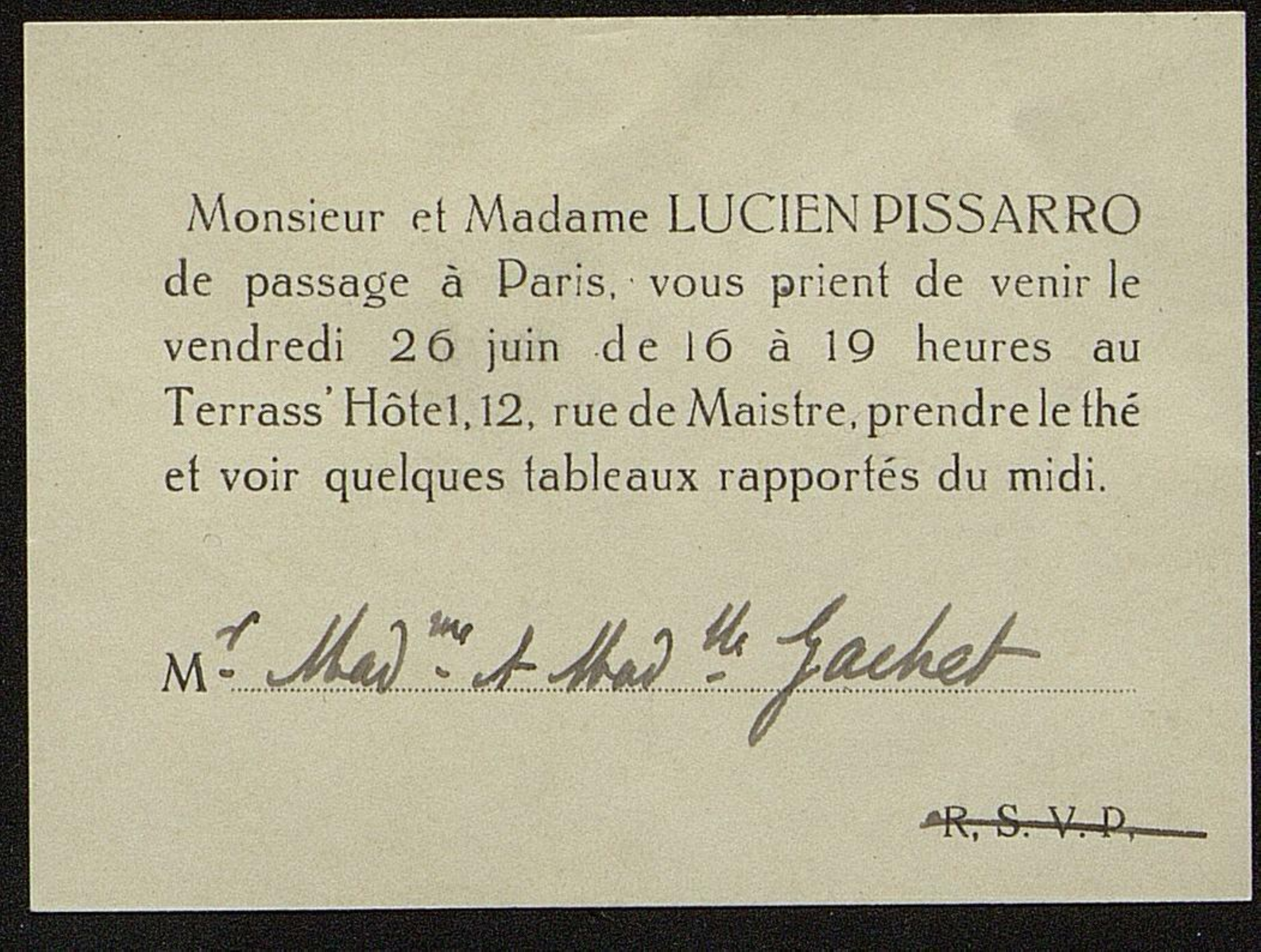}} \\
    \subfloat{\includegraphics[width=\textwidth, height=0.3\textheight, keepaspectratio]{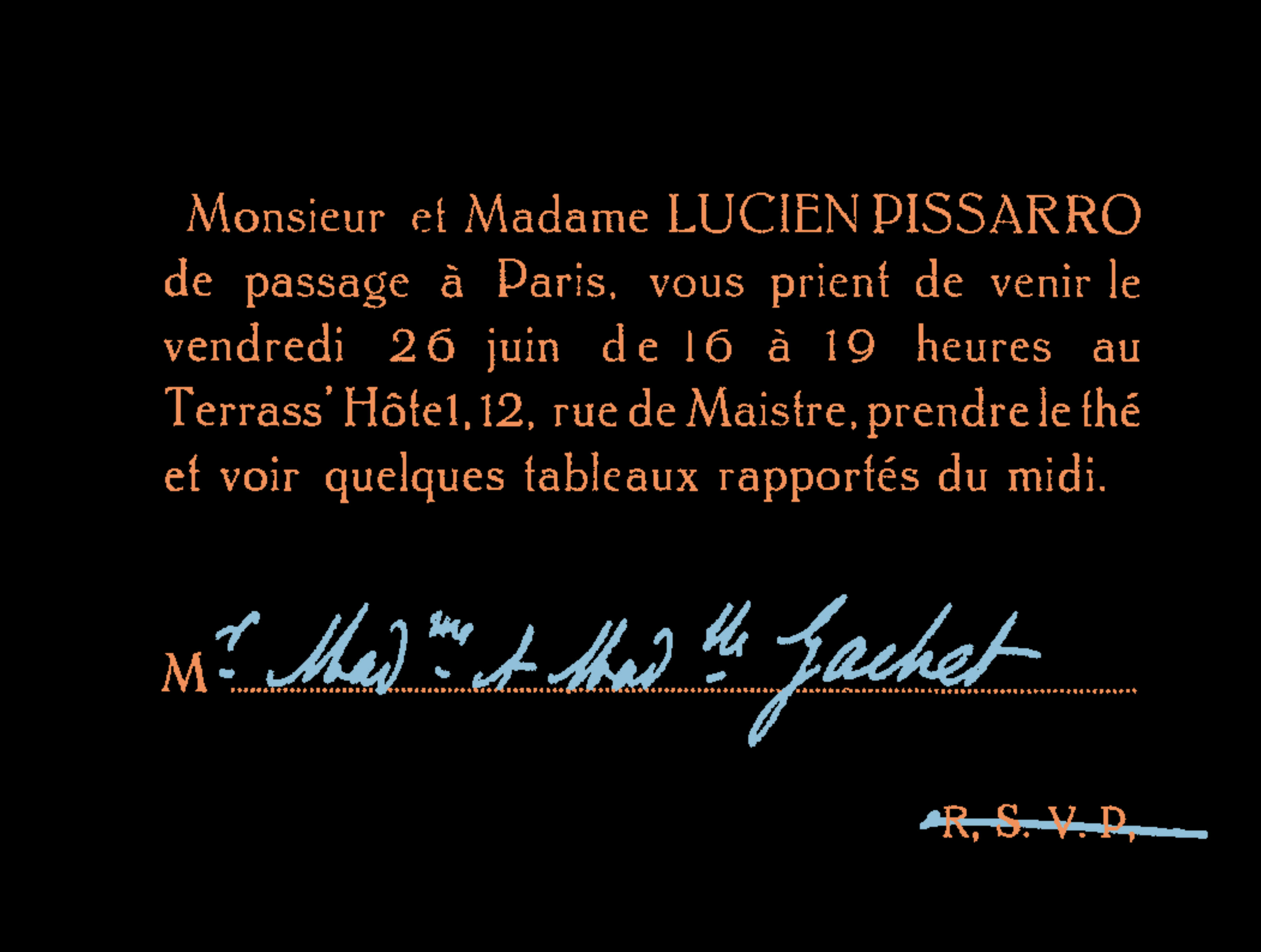}} \\
    \subfloat{\includegraphics[width=\textwidth, height=0.3\textheight, keepaspectratio]{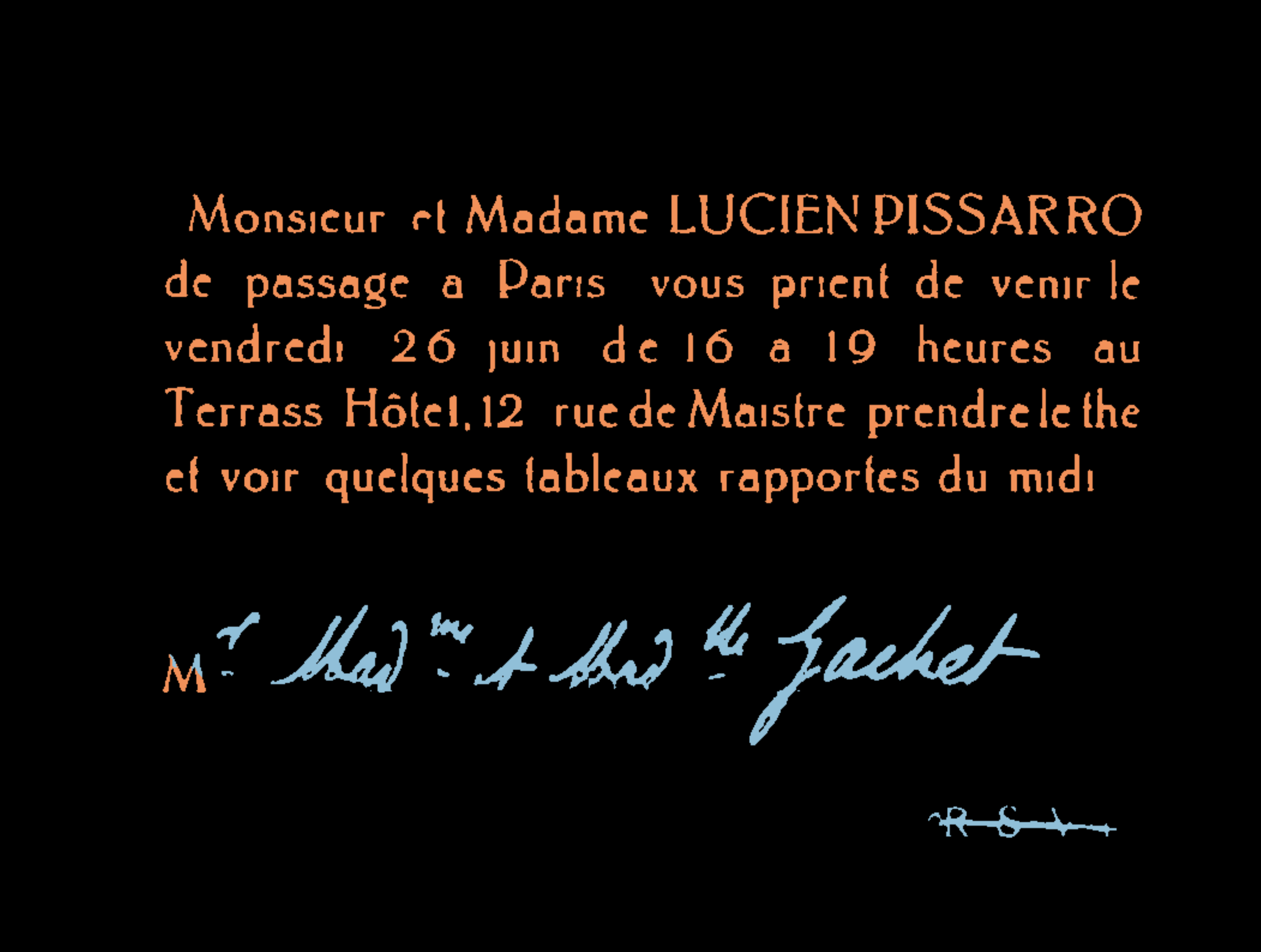}}
    \caption{Demonstration of our approach applied to an out-of-domain sample. Bottom: RGB image, Middle: Ground truth segmentation, Top: Prediction of TransUNet trained on synthetic data generated with our approach. The prediction is done on image patches that are then reassembled into the segmentation of the full image}
    \label{fig:qualitative_4}
  \end{figure*}

\end{document}